\documentclass[10pt, a4paper]{article}

\usepackage[]{lrec-coling2024} 
\usepackage{amsmath}
\usepackage{multirow}
\usepackage{booktabs}       
\usepackage{makecell}
\usepackage{subcaption}  
\usepackage{siunitx}  
\usepackage{algorithm}
\usepackage{algpseudocode}

\algnewcommand\algorithmicforeach{\textbf{for each}}
\algdef{S}[FOR]{ForEach}[1]{\algorithmicforeach\ #1\ \algorithmicdo}

\title{Project MOSLA: \\ Recording Every Moment of Second Language Acquisition}

\name{Masato Hagiwara \hspace{4em} Joshua Tanner} 

\address{Octanove Labs \& Earth Species Project \hspace{3em} Mantra Inc. \hspace{5em} \\
         Seattle, WA, USA \hspace{7em} Tokyo, Japan \\
         masato@octanove.com \hspace{4.5em} josh@mantra.co.jp \\}

\abstract{
Second language acquisition (SLA) is a complex and dynamic process. Many SLA studies that have attempted to record and analyze this process have typically focused on a single modality (e.g., textual output of learners), covered only a short period of time, and/or lacked control (e.g., failed to capture every aspect of the learning process). In Project MOSLA (Moments of Second Language Acquisition), we have created a longitudinal, multimodal, multilingual, and controlled dataset by inviting participants to learn one of three target languages (Arabic, Spanish, and Chinese) from scratch over a span of two years, exclusively through online instruction, and recording every lesson using Zoom. The dataset is semi-automatically annotated with speaker/language IDs and transcripts by both human annotators and fine-tuned state-of-the-art speech models. Our experiments reveal linguistic insights into learners' proficiency development over time, as well as the potential for automatically detecting the areas of focus on the screen purely from the unannotated multimodal data. Our dataset is freely available for research purposes and can serve as a valuable resource for a wide range of applications, including but not limited to SLA, proficiency assessment, language and speech processing, pedagogy, and multimodal learning analytics. \newline 
}

\begin{document}

\maketitleabstract

\section{Introduction}

The acquisition of a second language is a complex and dynamic process characterized by various milestones and challenges that learners encounter along their journey. Many studies have attempted to record the learning process, although most studies are unimodal (e.g., capturing only the textual output of learners, \citealp{geertzen2014automatic}), cover only a short period (e.g., containing snapshots of learner's progress, \citealp{settles2018second}), and/or are limited in control (e.g., not capturing every aspect of the learning process, \citealp{stasaski2020cima}). It has long been recognized that multimodal, longitudinal interaction is a crucial factor in SLA~\cite{hampel2012use}. 

In order to shed light on the complex and dynamic nature of the SLA process, in Project MOSLA (Moments of Second Language Acquisition), we created a longitudinal, multimodal, multilingual, and controlled dataset by inviting participants to learn a new language from scratch solely through online instruction over a span of two years and documenting every lesson using Zoom. This dataset, comprising over 250 hours of recorded lessons, captures the rich and nuanced aspects of language learning, including verbal and non-verbal communication, the use of teaching materials, student-teacher interactions, and the evolving proficiency of learners. Notably, the MOSLA dataset encompasses a diverse set of target languages---Arabic, Spanish, and Chinese---including two languages that employ non-Latin alphabets, highlighting the dataset's unique cross-linguistic scope. 

\begin{figure}[!t]
\begin{center}
\includegraphics[scale=0.36]{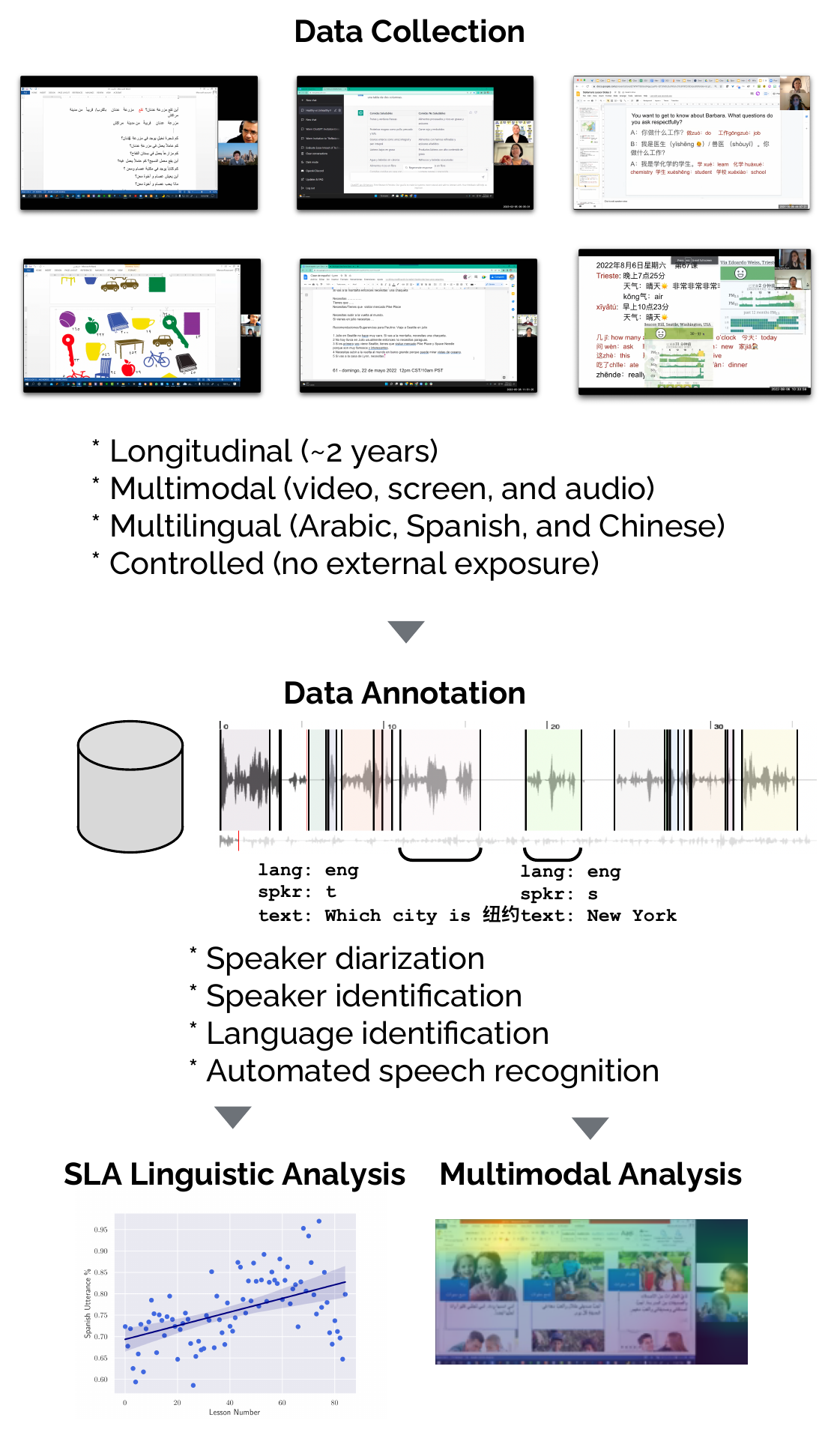} 
\caption{Overview of Project MOSLA}
\label{fig_overview}
\end{center}
\end{figure}

To enhance the dataset's utility, we semi-automatically annotated all the utterances in the recorded audio with start and end offsets, speaker and language IDs, and transcripts. This annotation was accomplished using human annotators and state-of-the-art machine learning models for speaker diarization, speaker and language identification, and automatic speech recognition. The resulting metadata offers valuable insights into the distribution of speech and speaker identities throughout the learning process, as well as transcriptions of spoken content. 

In this paper, we provide an overview of the creation, annotation, analysis, and applications of the MOSLA dataset. We begin by describing our data collection method and then discuss the process of human and machine annotation. We empirically demonstrate that fine-tuning state-of-the-art speech models with a small amount of human-annotated data results in substantial improvements in speaker and language identification, as well as speech recognition performance. Additionally, we show that our data can reveal linguistic insights into the learners' acquisition process of the target language, such as the percentage of non-English utterances and lexical diversity. Furthermore, we demonstrate that, through the use of deep neural network models, we can determine where on the screen the teacher and the learner are focusing, solely from the unannotated multimodal video and audio data. The MOSLA dataset represents a significant contribution to the field of SLA research, providing a rich source of data for investigating the factors influencing language learning outcomes, the role of multimodal cues in the acquisition process, and the development of innovative educational tools.

The MOSLA dataset is freely available for research and non-commercial purposes, ensuring that it can benefit the broader academic community and contribute to advancements in the field of second language acquisition. It can be accessed here: \href{https://www.octanove.com/mosla.html}{https://www.octanove.com/mosla.html}. 

\section{Related Work}

In the field of SLA, there have been many studies that aimed to record and analyze the learning process, providing valuable insights into language learning. However, many of these studies have limited temporal coverage, typically spanning only several months~\cite{vercellotti2015development,saito2017video-based}. Duolingo publishes the Second Language Acquisition Modeling (SLAM) dataset~\cite{settles2018second}, which contains learner production in their target language. However, the data covers only a 30-day period, offering a relatively short-term perspective on language acquisition. The CIMA dataset~\cite{stasaski2020cima} contains tutor-learner interaction data during language learning, but it lacks multimodal and longitudinal characteristics. Similarly, the Teacher-Student Chatroom Corpus~\cite{caines2020teacher} collected textual interactions between teachers and students during online English teaching but also lacks multimodal and longitudinal aspects.

In the realm of grammatical error correction (GEC), there is a substantial body of research~\cite{bryant2023grammatical} but relatively few GEC corpora focus on longitudinal learning. One noteworthy exception is the EFCamDat corpus~\cite{geertzen2014automatic}, one of the largest GEC corpora, with a collection period spanning a few years. However, only a few of its users participated over the entire duration, with many starting or ending their learning outside the collection period. 

In other domains of learning analytics, \citet{kubat2007totalrecall} collected two years' worth of multimodal data on first language development through the Human Speechome Project (HSP)~\cite{roy2006human}, primarily focusing on first language acquisition and involving data from a single individual. \citet{demszky2023ncte} collected and analyzed transcripts of teacher-student discourse in elementary math classrooms. 

MOSLA is closely related to the field of multimodal learning analytics (MMLA)~\cite{mu2020multimodal} and web-based language learning (WBLL)~\cite{CongLem2018WebBasedLL}. For example, \citet{donnelly2017words} analyzed classroom audio recordings, and \citet{monkaresi2017automated} examined facial expressions as part of the learning analytics process.

The MOSLA dataset holds the potential to be valuable for various applications, including assessment~\cite{settles2020machine}, proficiency estimation~\cite{vajjala2018experiments}, knowledge tracing~\cite{piech2015deep}, grammatical error correction~\cite{bryant2023grammatical}, automated assessment of speaking proficiency~\cite{fan2020Assessing}, and optimization of pedagogical approaches~\cite{lepper2002wisdom}, among others. Its longitudinal, multimodal nature makes it a unique resource for studying the complexities of the SLA process.

\section{Data Collection}

Data collection took place between February 2021 and February 2023. The teacher and the learner had weekly language instruction over zoom. Specifically,

\begin{table}
    \centering
    \begin{tabular}{crr} \toprule
                &  \# Videos & Total duration \\ \midrule
        Arabic  &  95 & 102 hrs \\
        Spanish &  85 & 84 hrs \\
        Chinese &  84 & 84 hrs \\ \bottomrule
    \end{tabular}
    \caption{Raw Statistics of Collected Data by Course}
    \label{tab:my_label}
\end{table}

\begin{itemize}
    \item A learner (a complete beginner) and a teacher have a private lesson per week online (e.g., on Zoom) for at least two years.
    \item Every lesson is recorded (video, audio, and screen share).
    \item The learner is not allowed to learn the target language outside of these lessons.
    \item All the materials the learner is exposed to are recorded (e.g., via screen share).
\end{itemize}
All the learners in this study were already proficient in two or more languages (their L1 and L2, typically English) before the study started and are generally highly motivated individuals. Below is additional information on the individual courses:

\begin{itemize}
    \item \texttt{ara}: Arabic (Modern Standard Arabic)
    \begin{itemize}
        \item Teacher L1: Levantine Arabic
        \item Learner L1: Japanese
        \item Learner L2s: English, Mandarin Chinese
        \item Learner Age: 35-44
        \item Learner Gender: Male
    \end{itemize}
    \item \texttt{spa}: Spanish (Latin American)
    \begin{itemize}
        \item Teacher L1: Spanish (Latin American)
        \item Learner L1: Mandarin Chinese
        \item Learner L2s: English, Japanese
        \item Learner Age: 35-44
        \item Learner Gender: Female
    \end{itemize}
    \item \texttt{zho}: Mandarin Chinese
    \begin{itemize}
        \item Teacher L1: Mandarin Chinese
        \item Learner L1: Spanish (Latin American)
        \item Learner L2s: English, Italian, German
        \item Learner Age: 25-34
        \item Learner Gender: Female
    \end{itemize}
    
\end{itemize}

All the teachers have a minimum of five years of professional experience teaching the target language. Additionally, all the participants are fluent in English, and the teaching instructions were conducted in English, at least initially. The study did not impose restrictions on the teaching methods employed by the instructors; they were free to use their preferred approaches. However, instructors were advised not to use copyrighted materials, such as textbooks and online courses, as-is, unless used in a supplementary capacity. As mentioned earlier, learners were prohibited from learning the target language outside of this study and were not assigned any explicit tasks beyond the classroom. Nevertheless, they were encouraged to review the recorded lesson videos for self-assessment purposes.

All the teaching was conducted via Zoom, and the video and audio were recorded using its standard recording functionality under the default settings. Participants used their own preferred devices for recording audio and video, which means that there was no quality control in regard to the devices.

\section{Data Annotation}

\begin{table*}[t]
  \centering
  \setlength{\tabcolsep}{10pt}
    \begin{tabular}{ c c c c c c c } 
    \toprule
     \thead{Language }& \thead{Source} & \thead{Total\\Duration} & \thead{Utterance\\Duration} & \thead{Utterance\\Count} & \thead{Target\\Language} & \thead{Student\\Utterance} \\
    \midrule
    \multirow{2}{4em}{Arabic} & \small Human & 3.0 hrs & 2.6 hrs & 2,330 & 82\% & 50\% \\ 
    & \small Machine &  101.5 hrs & 73.9 hrs & 80,441 & 81\% & 57\% \\ \midrule
    \multirow{2}{4em}{Spanish} & \small Human & 2.5 hrs & 2.2 hrs & 1,006 & 85\% & 52\% \\ 
    & \small Machine & 83.7 hrs & 61.1 hrs & 62,980 & 82\% & 50\% \\ 
        \midrule
    \multirow{2}{4em}{Chinese} & \small Human & 4.0 hrs & 3.3 hrs & 4,375 & 66\% & 24\% \\ 
    & \small Machine & 84.4 hrs & 65.6 hrs & 58,917 & 72\% & 33\% \\
    \bottomrule
    \end{tabular}
  \caption{Annotation Data}
  \label{tab:annotation}
\end{table*}

In addition to the video data for each lesson, MOSLA includes two sets of annotations containing information about the speech of the student and teacher: a smaller human-annotated set, and a complete machine-annotated set. Data from the human-annotated set is used to train machine learning models as shown in Figure~\ref{fig_pipeline}, which generate a complete set of data for all lessons. We release all models trained this way for use in future research. 

\subsection{Human Annotation}
We employ a bilingual annotator for each language pair, such that the annotator speaks both English and the language being learned. Annotation is done on five minute samples, which are selected as follows: we perform an independent random trial with a 5\% chance to succeed for each possible sample, and keep up to one sample per lesson\footnote{As an exception to this, a small number of Chinese lessons are slightly over-annotated.}. The first and last segment of each file are excluded from possible selection, as these often consist of technical setup or greetings instead of language education content. 

\begin{figure}[!t]
\begin{center}
\includegraphics[scale=0.26]{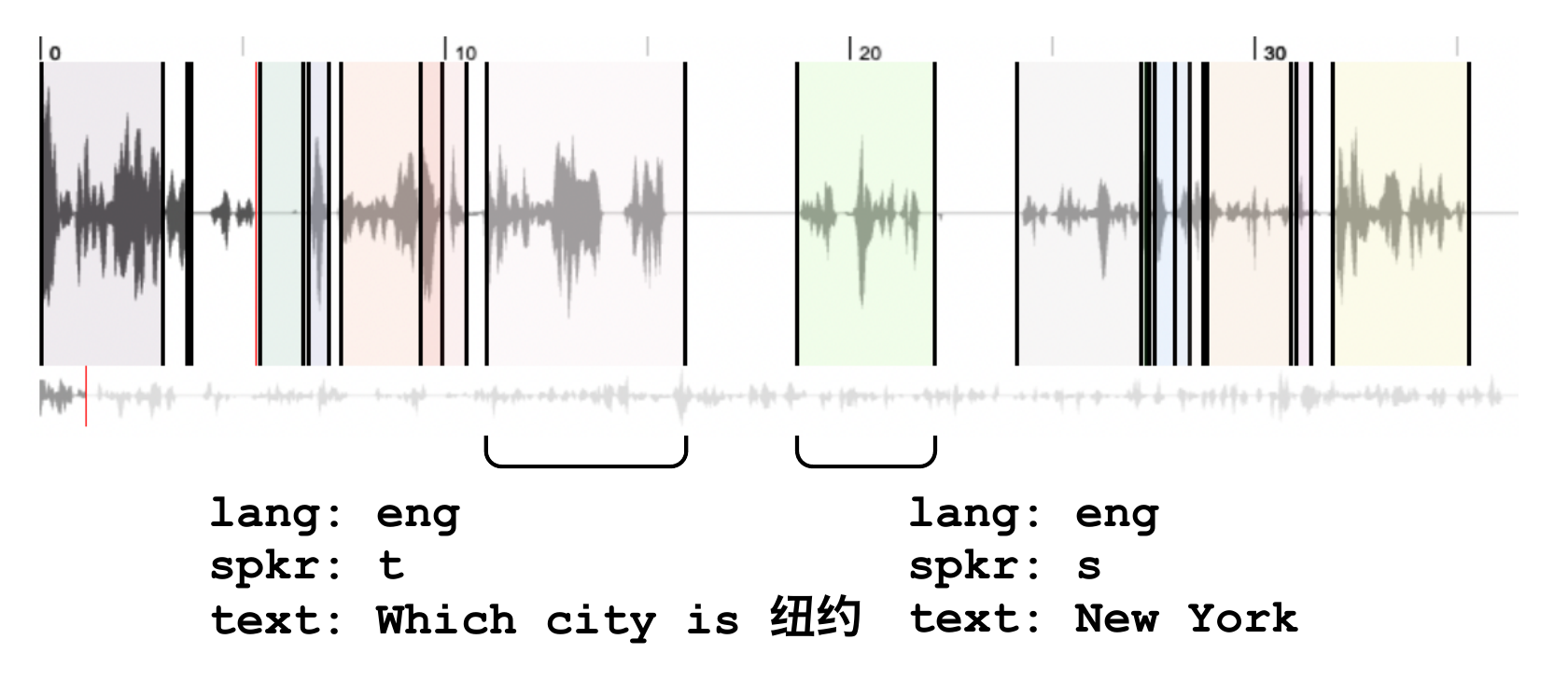} 
\caption{Example Annotation}
\label{fig_annotation}
\end{center}
\end{figure}

We use Hachiue~\cite{hachiue} for annotation, as it provides an easy to use web interface which allows annotators to mark arbitrary sections of the file as utterances and attach data to them. Annotators were instructed to create segments for distinct utterances from each speaker which include a speaker label (teacher, student or other), a label for the dominant language of the utterance (there are a number of code-switched utterances containing multiple languages), and a literal transcription of the speech. An example of segment annotations is shown in Figure~\ref{fig_annotation}. 

\subsection{Machine Annotation}

\begin{figure}[!t]
\begin{center}
\includegraphics[scale=0.41]{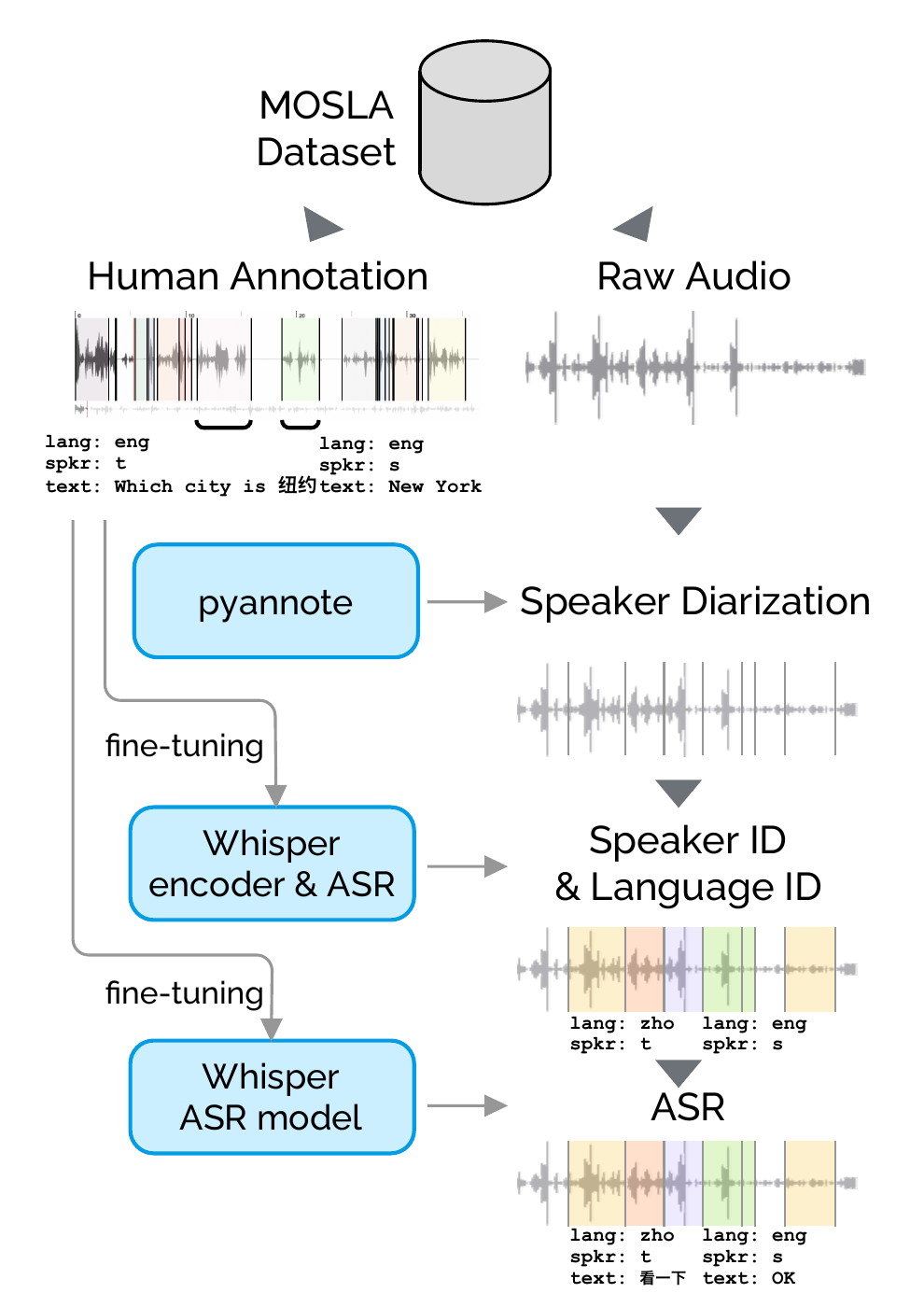} 
\caption{Overview of the annotation pipeline}
\label{fig_pipeline}
\end{center}
\end{figure}

Using the human annotation data, we train and evaluate machine learning models to perform each task necessary for annotation: diarization, speaker and language classification, and automatic speech recognition. These models are then combined into a machine annotation pipeline (Figure~\ref{fig_pipeline}) that we use to generate a set of annotations for the full duration of every lesson. The structure of the machine annotations are identical to the human annotations. 

We randomly select 20\% of annotated segments for each language to use as evaluation data, and use the remainder of the annotated data as training data. Summary statistics about both sets of annotation data can be seen in Table~\ref{tab:annotation}.

\subsubsection{Diarization}
We use Pyannote~\cite{Bredin23, Plaquet23} for speaker diarization. We experiment with both fine-tuning the speaker embeddings and segmentation model and with fine-tuning the diarization hyperparameters, but ultimately find the default pipeline to be the most performant. Note that because we perform speaker classification as a separate supervised task, we do not actually use the speaker clustering labels from the diarization pipeline, and care only about speech segmentation. This means that we could perform only voice activity detection (VAD) instead of full speaker diarization, but we found that diarization outperformed plain VAD for speech segmentation. Diarization also has the benefit of allowing us to process overlapping utterances from different speakers. 

Segmentation F1-score for our system can be seen in Table~\ref{tab:diarization}. While Pyannote's diarization model performed well enough to produce useful results, it is also the weakest component of our pipeline, likely due to background noise and fluctuating audio quality in the lesson recordings. We experimented with other diarization models such as those from the NeMo toolkit~\cite{harper2019nemo}, but were unable to find any which outperformed Pyannote.  

\begin{table}[ht]
    \centering
    \begin{tabular}{c c c} \toprule
                &  Diarization & VAD Only\\ \midrule
        Arabic &  79.6 & 79.0 \\
        Spanish &  69.4 & 66.6 \\
        Chinese  &  86.1 & 84.6 \\
        \bottomrule
    \end{tabular}
    \caption[Segmentation F1]{Segmentation F1 score, computed as the harmonic mean of purity and coverage\footnotemark}
    \label{tab:diarization}
\end{table}
\footnotetext{For details on purity, coverage and Segmentation F1 computation, see the \href{https://pyannote.github.io/pyannote-metrics/reference.html\#segmentation}{Pyannote metrics documentation}.}

\subsubsection{Utterance Classification}
We treat language and speaker identification as supervised classification tasks, where the input is the audio of a single utterance and the output is a label representing the dominant language in the utterance or the speaker of the utterance, respectively. 

We primarily experiment with Whisper~\cite{radford2022whisper} for these tasks, as it is known to perform well not only on automated speech recognition (ASR) but also on related speech and sound detection tasks~\cite{gong2023whisper-at}. We found that Whisper (the \texttt{whisper-large-v2} model) performs quite well when fine-tuned on our data. For speaker classification, we remove Whisper's autoregressive decoder component and replace it with a simple linear classification head with one hidden layer, such that the output of Whisper's encoder is fed directly into the classifier. As can be seen in Table~\ref{tab:speakid} and the classification row for Table~\ref{tab:langid}, this configuration performs fairly well on our data. 

\begin{table}[ht]
\centering
    \begin{tabular}{l c c c} \toprule
        & Arabic & Spanish & Chinese \\
        \midrule
        Classification & 90\% & 92\% & 95\% \\
        \bottomrule
    \end{tabular}
    \caption{Speaker identification accuracy}
    \label{tab:speakid}
\end{table}
\noindent We try the same classifier configuration for language identification, but find that our best performance comes from using the standard Whisper architecture, including decoder, fine-tuned on our data. That is, we use our ASR model for language identification by taking a single decoding step and selecting the most likely language token representing either English or the target language. We hypothesize that this performance gap exists because Whisper is already trained to output language tokens, and consequently has learned how to perform language identification using parameters in its decoder. 

\begin{table}[ht]
\centering
    \begin{tabular}{l c c c} \toprule
        & Arabic & Spanish & Chinese \\
        \midrule
        \textit{whisper-large-v2} & 46\% & 59\% & 76\%  \\ 
        ASR fine-tuned & 95\% & 95\% & 92\% \\
        Classification  & 95\% & 89\% & 90\% \\
        \bottomrule
    \end{tabular}
    \caption{Language identification accuracy}
    \label{tab:langid}
\end{table}

\subsubsection{Automatic Speech Recognition}
We also use Whisper for automatic speech recognition (ASR), finding once again that fine-tuning on our annotated data substantially improves performance. For both training and evaluation, the input in all cases is a single utterance as annotated by our human annotators, with the annotated speech as gold output labels. Note that we also provide the language of the utterance to the model by forcing the first decoded token to be the language token representing the utterance's dominant language. We use human-annotated gold language labels when training and evaluating our ASR models. Both classification and ASR models were fine-tuned for three epochs with a batch size of eight and a learning rate of $1 \times 10^{-6}$, using the cross-entropy loss. We measure ASR performance with character error rate (CER), in part because there is no standard way to calculate word error rate (WER) for languages without spaces like Chinese. Character error rate can be thought of as a measurement of the edit distance between the output of the model and the reference transcription. That is, given a reference of length $N$ characters and model output which can be transformed into this reference with $S$ substitutions, $D$ deletions and $I$ deletions, CER is computed as:
\begin{align}
    \mathit{CER} = \frac{S+D+I}{N}
\end{align}
ASR model performance can be seen in Table~\ref{tab:asr}. Fine-tuning the model improves performance on all languages, but most dramatically for Arabic and Chinese, where the error rates after fine-tuning are nearly half of the original. We speculate that Whisper may have benefited more from fine-tuning in these two languages because it was weaker in them to begin with: Whisper's reported ASR performance on Arabic and Chinese was substantially worse than Spanish in the original work \cite{radford2022whisper}.

\begin{table}[ht]
\centering
    \begin{tabular}{l c c c} \toprule
        & Arabic & Spanish & Chinese \\
        \midrule
        \textit{whisper-large-v2} & 60\% & 33\% & 32\% \\ 
        ASR fine-tuned & 25\% & 28\% & 17\%  \\
        \bottomrule
    \end{tabular}
    \caption{CER on each language for ASR models. Punctuation and Arabic diacritics are excluded for all CER computation.  }
    \label{tab:asr}
\end{table}

\subsubsection{Pipeline Scoring \& Error Propagation}
Scores in the previous sections are computed by comparing model outputs to human outputs for each human-annotated utterance. However, when running the machine annotation pipeline there is no guarantee that output from diarization or other steps will be correct, and consequently we can expect some degree of error propagation to later tasks in the pipeline. In particular, errors in speech segmentation are potentially damaging to all other tasks, and errors in language classification could lead to worse ASR output because the utterance language is used to bias the ASR model's output. 

Because diarization output will not line up perfectly with human annotated utterances,  we compute metrics per five minute human-annotated segment instead of per utterance in order to accurately gauge the performance of our pipeline. CER is computed by concatenating the speech in all utterances output by the pipeline and comparing it to the concatenation of all human-annotated utterance text. For speaker and language identification, we compute the identification error rate (IER) for each. IER can be thought of as a measurement of the percentage of the total duration that is classified incorrectly in some way, and is calculated as:
\begin{align}
    \mathit{IER} &= \frac{f + m + c}{t}
\end{align}
Where $f$ is the duration of false positives (non-speech incorrectly identified as speech), $m$ is the duration of missed speech (speech incorrectly identified as non-speech), $c$ is the duration of correctly identified speech assigned the wrong classification label, and $t$ is the total duration. Note that because $f$ and $m$ both depend exclusively on the performance of the model identifying speech, IER is particularly sensitive to diarization performance.

\begin{table}[ht]
\centering
\small

    \begin{tabular}{l l c c c} \toprule
        & & Arabic & Spanish & Chinese \\
        \midrule
        \multirow{2}{4em}{Spk ID} & Gold Seg & 5\% & 10\% & 4\% \\
        & Pipeline & 24\% & 27\% & 17\% \\
        \midrule
        \multirow{2}{4em}{Lang ID} & Gold Seg & 4\% & 3\% & 5\% \\
        & Pipeline & 24\% & 23\% & 21\% \\
        \midrule
        \multirow{3}{4em}{ASR} & Gold Seg & 23\% & 27\% & 16\% \\
        & $-$\scriptsize Gold Lang & 28\% & 27\% & 17\% \\
        & Pipeline & 34\% & 33\% & 31\% \\
        \bottomrule
    \end{tabular}
    \caption{Error rates for pipeline components: CER for ASR and IER for classification}
    \label{tab:error_prop}
\end{table}

\noindent In Table~\ref{tab:error_prop}, we present the performance of our pipeline components using human-annotated gold speech segmentation, and pipeline diarization. We also include ASR with gold speech segmentation but pipeline language identification. As we can see from these results, errors in diarization have a substantial effect on the performance of downstream tasks. Precise start and end times for utterances are arguably not necessary for downstream analysis focused on speech content, suggesting that the increase in IER for classification tasks may not matter in some cases, but errors in diarization also lead to an average increase in CER of approximately 10\% for ASR. We leave speech segmentation approaches which are more resilient to issues such as variable audio quality to future work.

\section{Experiments}

\subsection{Linguistic Analysis}
\begin{figure*}[ht]
\centering
\begin{subfigure}{.5\textwidth}
  \centering
  \includegraphics[width=\linewidth]{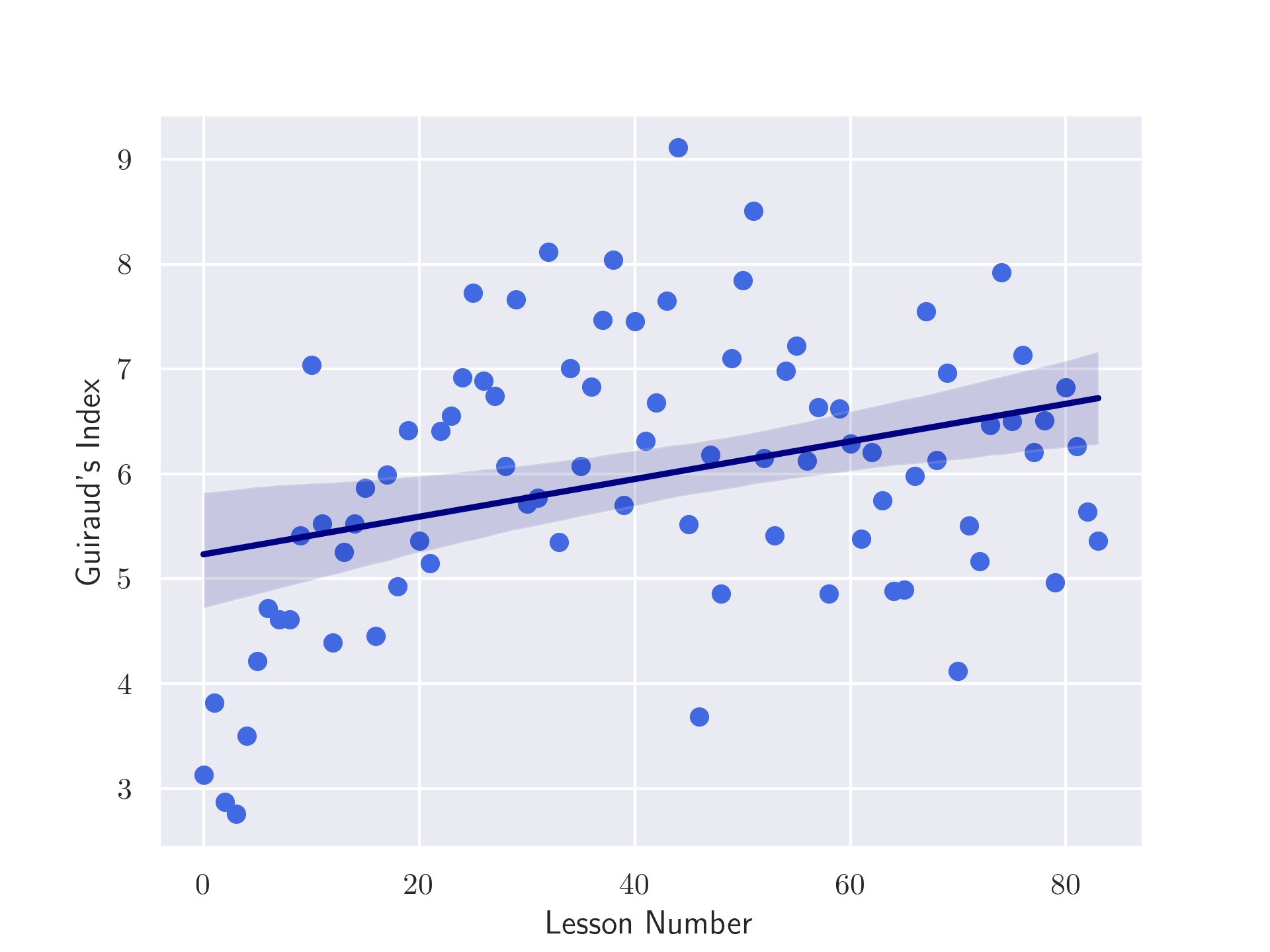}
    \caption{Student}
  \label{fig:sub1}
\end{subfigure}%
\begin{subfigure}{.5\textwidth}
  \centering
  \includegraphics[width=\linewidth]{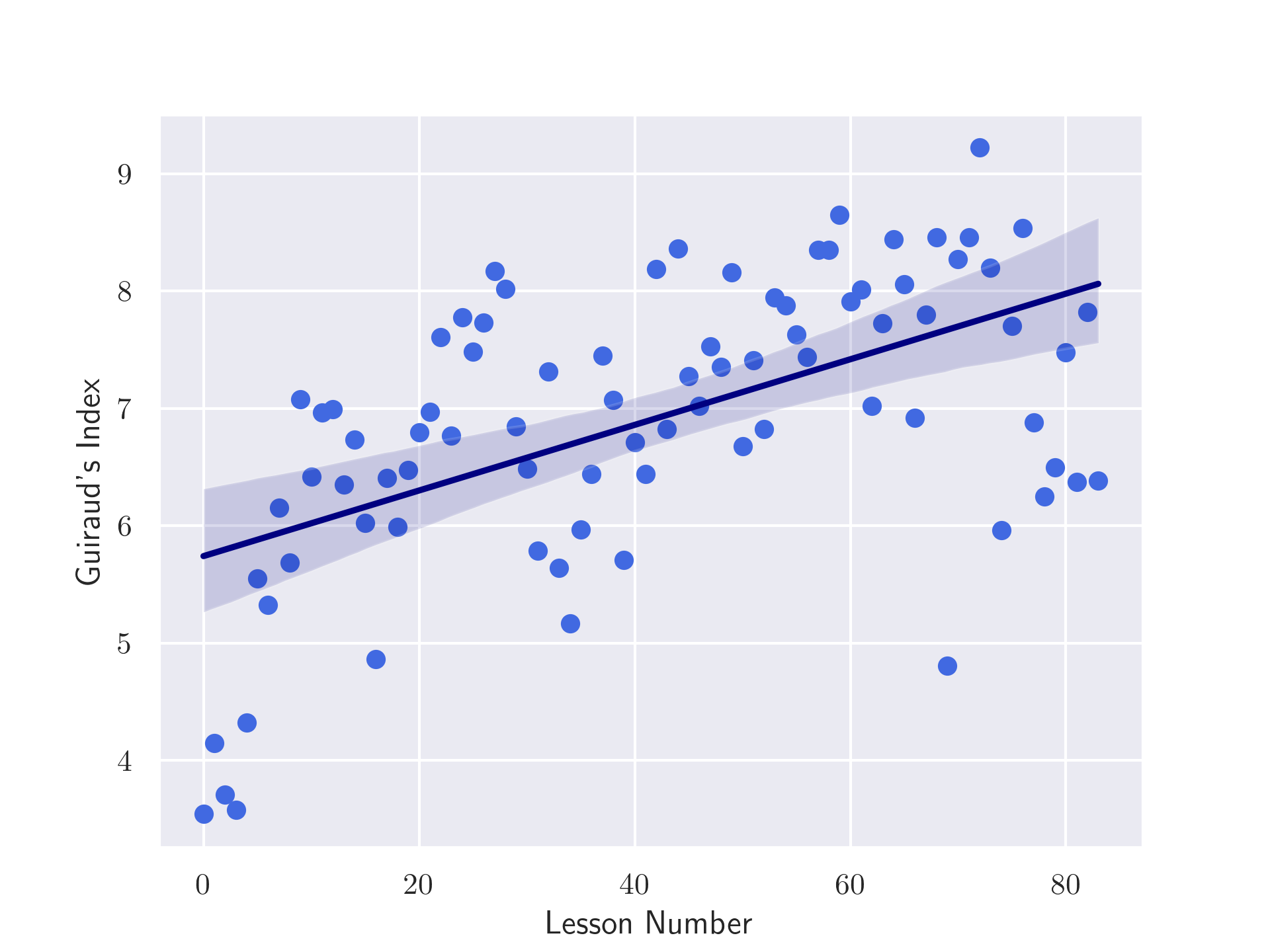}
    \caption{Teacher}
  \label{fig:sub2}
\end{subfigure}
\caption{Guiraud's index for the Chinese student and teacher}
\label{fig:viz_guiraud}
\end{figure*}

To demonstrate the kind of analysis that our data can be used for, we compute summary statistics to track changes in the learner and teacher's speech over time. We use a mix of human and machine-annotated data for this, using human data where available and machine-annotated data otherwise. 

\begin{figure}[H]
    \centering
    \includegraphics[width=\linewidth]{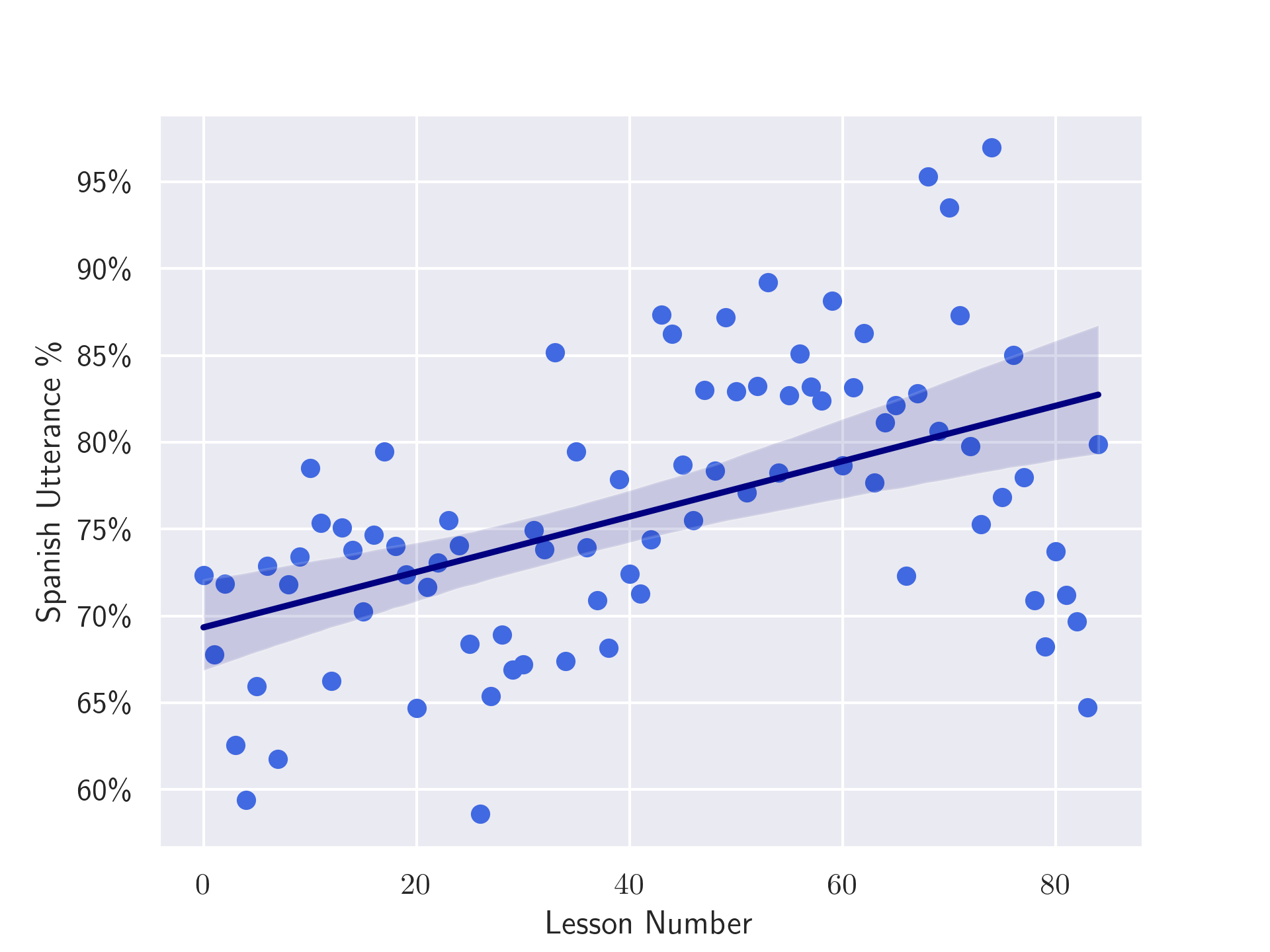}
    \caption{Student Spanish utterance \%}
    \label{fig:viz_nonenglish}
\end{figure}

We begin by examining the percentage of utterances made in the target language by both the teacher and student in each lesson. This is important both because listening and speaking practice are critical to language acquisition, and because for students the degree of target language use in a learning context has been linked to proficiency in that language~\cite{turnbull2009first, carranza1995multilevel}. We find that the percentage of target language utterances consistently increases over time for both the student and teacher: Spearman's $\rho$ for the correlation between lesson number and \% of target language utterances ranged from 0.32 to 0.73, with all $p$ values $< 0.01$. Data for the Spanish student can be seen in Figure~\ref{fig:viz_nonenglish}. All figures presented in this section are linear regressions with 95\% confidence intervals.

\noindent Next, we look at metrics designed to measure lexical diversity in speech, as they have been shown to correlate with assessments of learner ability~\cite{engber1995relationship} and can grow over time for language learners~\cite{hsieh2016exploratory}. For computing these metrics, we tokenize Spanish and Chinese using spaCy (which internally uses pkuseg for Chinese)~\cite{honnibal2020spacy, pkuseg2019luo}, and Arabic with CAMeL tools~\cite{obeid2020camel}. Token data is then cleaned by removing tokens consisting of punctuation, numbers, whitespace and stop words. 

Token-type ratio (TTR) is one measure of lexical diversity which is commonly used in linguistics research~\cite{thomas2005type}. TTR is calculated as the number of unique tokens (words) divided by the number of total tokens, and ideally would be expected to increase over time as the learner's vocabulary expands and the teacher moves on to using more complex language. However, TTR has been shown to be unstable in some circumstances, such as when there is substantial variance in the total number of tokens~\cite{van2007comparing}. This instability is mirrored in our results: while TTR grows over time for some students and teachers in some languages, correlations are often weak or have $p$ values substantially higher than $0.05$ suggesting no correlation at all. 

Some alternatives to TTR have been proposed to address its shortcomings. In particular, we look at Guiraud's index \cite{guiraud1954caracteres}, which mitigates the influence of total number of tokens by using the square root of the total token count as the denominator. We present standard TTR and Guiraud's index below as $\mathit{TTR}$ and $\mathit{TTR}_{guiraud}$ below, where $N$ is the total number of tokens and $V$ is the number of \textit{unique} tokens. 

\begin{align}
    \mathit{TTR} &= \frac{V}{N} &  \mathit{TTR}_{guiraud} &= \frac{V}{\sqrt{N}}
\end{align}
\begin{table}[ht]
\centering
\resizebox{\columnwidth}{!}{%
\small
    \begin{tabular}{l l c c c } \toprule
        Metric & & \small Arabic & Spanish & Chinese \\
        \midrule
        \multirow{2}{7em}{Target Lang \%} & Student & 0.58 & 0.48 & 0.46  \\
        & Teacher & 0.72 & 0.32 & 0.73 \\
        \midrule
        \multirow{2}{7em}{Guiraud's Index} & Student & 0.32 & 0.38 & 0.30 \\
        & Teacher & 0.37 & 0.55 & 0.53 \\
        \bottomrule
    \end{tabular}
    }
    \caption{Spearman's $\rho$ for correlation between summary statistics and lesson number. All correlations have $p < 0.01$.}
    \label{tab:summary_stats}
\end{table}

\noindent We find that Guiraud's, like \% of target language utterances, consistently increases over time (i.e. correlates with lesson number) for both students and teachers as can be seen in Table~\ref{tab:summary_stats}. Spearman's $\rho$ ranges from 0.30 to 0.55 with all $p$ values $<0.01$. Interestingly, this effect is measurably stronger for teachers, who had a mean $\rho$ of 0.48 as opposed to students' 0.33. A comparison of change in Guiraud's index for the Chinese student and teacher can be seen in Figure~\ref{fig:viz_guiraud}. Assuming that students made measurable progress over the course of their lessons and that teachers gradually increased the difficulty of lesson content, these results show that this progression is reflected in our data, and also speak to the suitability of Guiraud's index as a metric.

\subsection{Multimodal Analysis}

\begin{figure*}[!ht]
\begin{center}
\includegraphics[scale=0.3]{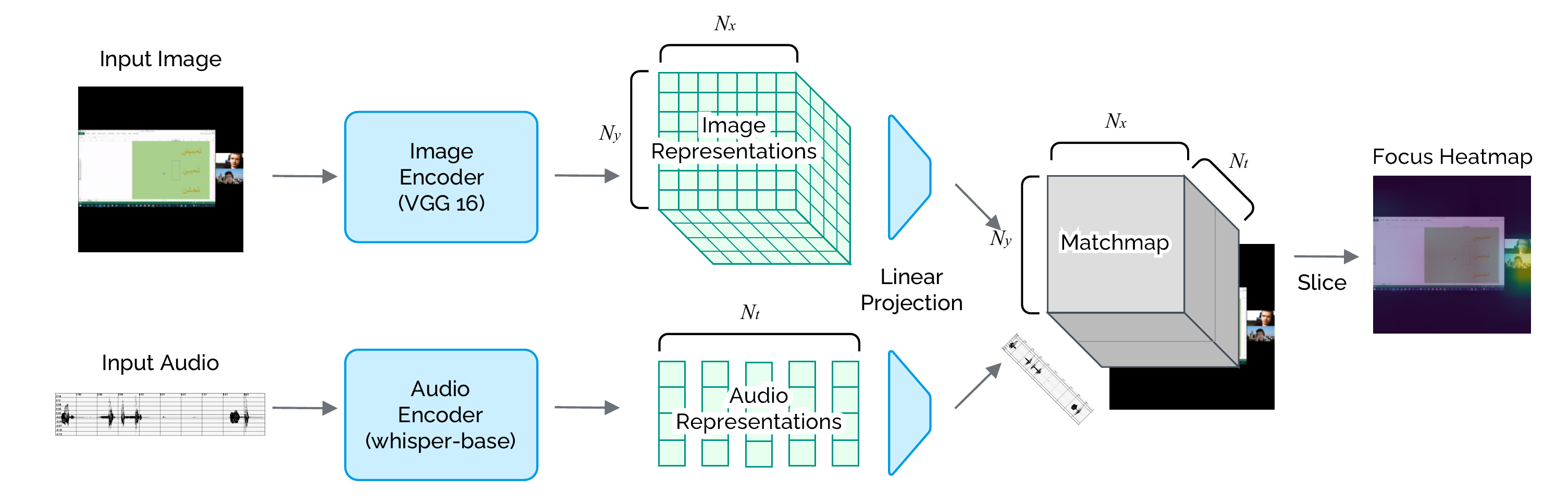} 
\caption{Overview of the Matchmap method}
\label{fig_matchmap}
\end{center}
\end{figure*}

We also illustrate how the rich multimodal data in the MOSLA dataset can be harnessed to gain insights into teacher and student behaviors using modern machine learning techniques.

Our objective here is to use machine learning techniques to determine the area of focus for both the teacher and the student on the screen, based solely on unannotated raw audio and video data.

Specifically, we use the Matchmap method, as described in \cite{harwath2018jointly}, to align the raw audio and the image in an unsupervised manner. The underlying principle of this method is that when parts of the input image and the audio co-occur frequently, it results in a high similarity score for that combination. The Matchmap method, as shown in Figure~\ref{fig_matchmap}, encodes an image and an audio clip using separate encoders, producing a grid or sequence of latent representations for each modality. Let $a_{t, u}$ be the $u$-th element of the audio representation vector $\mathbf{a}_t$ at time $t$, and $i_{x, y, u}$ be the $u$-th element of the image representation vector $\mathrm{i}_{x, y}$ at position $(x, y)$. After applying a linear projection layer ($f_a$ and $f_i$, respectively) to each modality, the method computes a three-dimensional matrix called Matchmap as:
\begin{equation}
    M_{x, y, t} = f_i(\mathbf{i})^T_{x, y} f_a(\mathbf{a})_t,
\end{equation}
which quantifies the degree of ``compatibility'' between the image at position $(x, y)$ and the audio at time $t$. Finally, the Matchmap matrix is aggregated to determine the overall similarity (referred to as SISA—Sum Image, Sum Audio) between a given image $I$ and audio $A$ instances using a simple arithmetic mean:
\begin{equation}
    S(I, A) = \frac{1}{N_x N_y N_t} \sum_{x, y, t} M_{x, y, t}
\end{equation}
where $N_x, N_y, N_t$ denote the width and height of the encoded image, and the length of the encoded audio sequence, respectively.

To learn the Matchmap matrix without the need for labels, we adopt a contrastive learning approach. This approach maximizes the similarity between true image-audio pairs $(I_i, A_i)$ while minimizing the similarity between randomly chosen ``imposter'' images $I^{imp}_i$ and audio $A^{imp}_i$. Specifically, the Matchmap method uses the following loss function as the learning objective:
\begin{align}
    L &= \sum_{i = 1}^{N_b} \left( \max(0, S(I_i, A^{imp}_i) - S(I_i, A_i) + \eta) \right. \nonumber \\
     &+ \left. \max(0, S(I^{imp}_i, A^i) - S(I_i, A_i) + \eta) \right)
\end{align}
where $N$ is the number of instance per batch. Imposter images and audio were created by randomly permutating the instances within each batch. We set $\eta = 1$ in our experiments.

We initially extracted 100 random 10-second chunks from each Arabic lesson video. Images were generated by calculating the average of all the frames within each chunk. In this experiment, we used the pretrained VGG16 model~\cite{simonyan2015very} before the last pooling layer for encoding images and the Whisper~\cite{radford2022whisper} encoder (base model) for audio. Audio representations were downsampled by averaging every 10-frame window, resulting in $7\times 7$ feature maps for images and 50 frames for audio. Both image and audio representations were then transformed using a 512-dimensional linear layer before computing the matchmap. The entire network, including the encoders and linear layers, was optimized using Adam with a learning rate of $1.0\times10^{-4}$ for 30 epochs, with a batch size of 32.

\begin{figure}[!t]
\begin{center}
\includegraphics[scale=0.28]{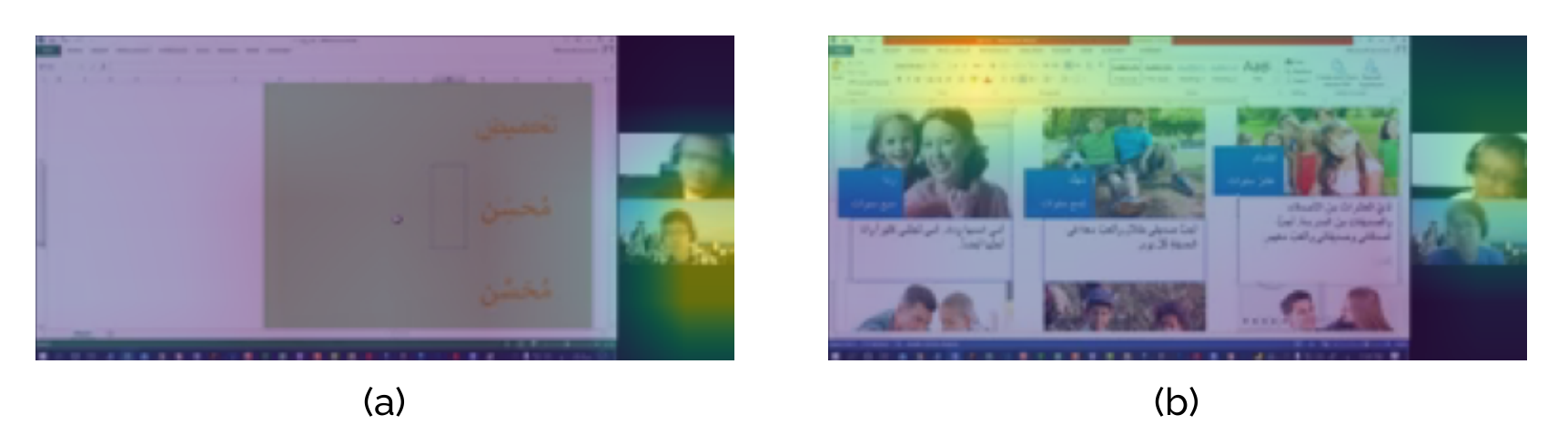} 
\caption{Example visualization of Matchmap}
\label{fig_matchmap_results}
\end{center}
\end{figure}

Figure~\ref{fig_matchmap_results} displays some examples of visualized matchmaps. These images were generated by slicing the computed matchmap at time $t$ when discourse is taking place, whether initiated by the teacher or student, and then overlaying it as a heatmap onto the original image. As can be seen in the figure, the matchmap highlights relevant parts of the input image, such as the speaker (a) and/or the learning content (b). While we have not conducted a formal evaluation of this model, these results suggest that similar multimodal analytics approaches may prove effective for tasks such as speaker diarization, automated speech recognition, and facial expression analysis.

\section{Conclusion}

In Project MOSLA (Moments of Second Language Acquisition), we address the complexity of SLA by creating a longitudinal, multimodal, multilingual, and controlled dataset that captures every moment of SLA learners' experiences through online instruction. With human and machine annotations generated using state-of-the-art speech models, the MOSLA dataset provides insights into the distribution of spoken language, speaker identities, and the content of spoken discourse. Our experiments highlight the potential of this resource in revealing target language usage and lexical development, as well as in identifying the areas of focus for both learners and educators during interactions. By offering open access to the MOSLA dataset for research and non-commercial purposes, we hope to inspire a wide array of studies, fostering a deeper understanding of the multifaceted nature of SLA and facilitating the development of more effective pedagogical approaches for second language learners.

\section{Ethical Considerations}
As it is difficult to imagine possible harms as a result of further research or technology built on a dataset about language acquisition, our primary ethical concerns relate to the fairness of compensation and exposure to risk for participants in the study. In regards to compensation: all participants---students, teachers, and annotators---were paid well above the minimum hourly wage in the country in which this research was conducted. 

We view risk to participants as consisting broadly of two categories: possible exposure of personally identifiable information (PII) relating to teachers or students, and possible appropriation of teaching material. Our primary mitigation against these risks is that access to MOSLA data will require consenting to a terms of use document which explicitly prohibits attempts to extract PII, appropriate teaching materials, redistribute the data, or otherwise use it for anything other than research. There is no explicit PII included anywhere in the data; our concern is only preventing the possibility of PII being inferred from conversation content in lessons. Furthermore, all participants knew from before their first lesson that they were being recorded with the intent of eventually publishing the data, had the option to withdraw at any point, and had and continue to have the right to request removal of any data, at any time, for any reason.

One other possible area of concern is copyrighted materials. In order to address this, teachers were asked to refrain from using copyrighted materials except in a supplementary capacity, and we are confident that any such usage included in the MOSLA lessons falls under fair use for teaching and research. 

Finally, in place of an IRB or equivalent institutional review board which we did not have access to, we had a third-party ethics review conducted by an external resesarcher with an extensive background in AI and data ethics.

\section{Acknowledgements}

We would like to thank Paulino Brener and Abdulwahed Danou for their participation as teachers and Xiaoling Mo for her participation as both teacher and learner. Participants are thanked in the acknowledgements at their request, and had the option to remain anonymous.

We would also like to thank Keisuke Sakaguchi and Kasumi Yamazaki for their valuable advice, which informed this research. Finally, we would like to thank Marius Miron for providing a third-party ethics review. 


\section*{Bibliographical References}\label{sec:reference}

\bibliographystyle{lrec-coling2024-natbib}
\bibliography{lrec-coling2024}

\end{document}